\pdfoutput=1

\documentclass[11pt]{article}

\usepackage[final]{acl}

\usepackage{times}
\usepackage{latexsym}
\usepackage{lipsum}
\usepackage{booktabs}
\usepackage[labelfont=bf]{caption}
\usepackage{multirow}
\usepackage{graphicx}
\usepackage{amssymb}
\usepackage{amsmath}
\usepackage{soul}

\usepackage[T1]{fontenc}
\usepackage[most]{tcolorbox}
\usepackage{listings}
\usepackage{xcolor} 
\lstdefinelanguage{json}{
    morestring=[b]",
    morecomment=[l]{//},
    morekeywords={true,false,null},
    stringstyle=\color{blue},
    keywordstyle=\color{purple},
    commentstyle=\color{gray},
    basicstyle=\ttfamily,
    showstringspaces=false
}

\usepackage[utf8]{inputenc}

\usepackage{microtype}

\usepackage{inconsolata}

\usepackage{graphicx} 
\usepackage{multirow}
\parskip=\medskipamount

\title{A Gold Standard Dataset and Evaluation Framework for Depression Detection and Explanation in Social Media using LLMs}

\author{
Prajval Bolegave, Pushpak Bhattacharya \\
Computer Science and Engineering, IIT Bombay, India \\
\texttt{\{prajval, pb\}@cse.iitb.ac.in}
}

\begin{document}
\maketitle

\begin{abstract}
Early detection of depression from online social media posts holds promise for providing timely mental health interventions. In this work, we present a high-quality, expert-annotated dataset of 1,017 social media posts labeled with depressive spans and mapped to 12 depression symptom categories. Unlike prior datasets that primarily offer coarse post-level labels \cite{cohan-etal-2018-smhd}, our dataset enables fine-grained evaluation of both model predictions and generated explanations.

We develop an evaluation framework that leverages this clinically grounded dataset to assess the faithfulness and quality of natural language explanations generated by large language models (LLMs). Through carefully designed prompting strategies, including zero-shot and few-shot approaches with domain-adapted examples, we evaluate state-of-the-art proprietary LLMs including GPT-4.1, Gemini 2.5 Pro, and Claude 3.7 Sonnet.

Our comprehensive empirical analysis reveals significant differences in how these models perform on clinical explanation tasks, with zero-shot and few-shot prompting. Our findings underscore the value of human expertise in guiding LLM behavior and offer a step toward safer, more transparent AI systems for psychological well-being.
\end{abstract}




\section{Introduction}

Mental health disorders, particularly depression, have become increasingly prevalent and are often expressed through social media platforms \cite{cohan-etal-2018-smhd, guntuku2017detecting}. The ubiquity of user-generated content has prompted researchers to explore language-based signals for early identification of psychological distress \cite{calvo2017natural, xu2023mental}. However, most existing machine learning models for depression detection operate on weakly-labeled data or binary classification tasks, limiting their clinical utility and interpretability \cite{chancellor2020methods}.

Further, the explainability of these models remains a critical bottleneck, especially when applied to sensitive domains like mental health. While large language models (LLMs) such as GPT-4 and Claude offer impressive fluency, their generated explanations often lack alignment with clinically grounded reasoning \cite{yang2023mentalllama, rajani2019explain}. This gap between linguistic coherence and domain-specific faithfulness raises important concerns for deploying AI in healthcare contexts \cite{doshi2017towards}.

To address these challenges, we introduce a method for constructing a high-quality, expert-annotated dataset for depression detection and explanation. Each post is manually labeled by licensed mental health professionals with two layers of annotation: (i) spans of text indicative of depressive symptoms, and (ii) corresponding mappings to one or more of twelve standard symptoms derived from the DSM-5 criteria (e.g., anhedonia, hopelessness, suicidal ideation) \cite{apa2013dsm}. This dual-layer annotation enables more granular supervision for both training and evaluation.

Our contributions are:
\begin{enumerate}
    \item \textbf{A gold-standard mental health dataset} consisting of $1,017$ social media posts annotated by clinical experts for both depressive spans and symptom categories, enabling rigorous evaluation of explainability in mental health prediction (Section \ref{dataset}).

    \item \textbf{An evaluation framework} leveraging the annotated dataset to assess the \textit{Faithfulness}, which is captured through Span-Recall, measuring the presence of expert-annotated depressive spans in model-generated explanation (weighted 30\%), \textit{Clinical relevance}, which is assessed using Symptom-F1, which evaluates alignment between predicted symptoms and ground truth labels (weighted 40\%), and \textit{Coherence Score}, reflecting the fluency and interpretability of the explanation (weighted 30\%). (Section \ref{eval_framework}). 
    
    \item \textbf{Prompting strategies} for explanation generation, including \textsc{Zero-Shot} and \textsc{Few-Shot} prompting with domain-adapted examples using advanced LLMs such as GPT-4.1, Gemini 2.5 pro, and Claude 3.7 Sonnet, to explore their impact on explanation quality in NLE tasks (Section \ref{method}, Appendix \ref{appendix prompts}).
    


    \item \textbf{Empirical comparison of proprietary LLMs} (GPT-4.1, Gemini 2.5 Pro, Claude 3.7) on expert-annotated mental health data, highlighting their relative strengths and weaknesses in clinical explanation tasks across different prompting settings. (Section \ref{results_analysis})

\end{enumerate}

\section{Related Work}

\subsection{Mental Health Datasets and Annotation Practices}

Annotated datasets have played a critical role in advancing NLP for mental health analysis. The SMHD corpus provided large-scale Reddit data labeled for multiple mental health conditions, enabling early detection through linguistic signals \cite{cohan-etal-2018-smhd}. EmoMent focused on emotion-annotated posts from South Asian populations, highlighting cultural nuances in mental health expression \cite{atapattu-etal-2022-emoment}. More recently, the IMHI dataset introduced by the MentaLLaMA framework offered expert-written explanations for multiple mental health tasks, supporting interpretable model development \cite{yang2023mentalllama}. However, these resources predominantly feature document-level annotations and lack token-level span highlights or symptom-specific mappings grounded in clinical criteria. In contrast, our dataset provides expert annotations at both the span and symptom level based on DSM-5 categories, enabling more granular and clinically faithful evaluation.

\subsection{Faithfulness and Explainability in LLMs}

Explainability has become a key concern in applying LLMs to sensitive domains such as mental health. The Mental-LLM benchmark evaluated models like GPT-3.5 and GPT-4 on diagnostic tasks and found limitations in zero-shot and few-shot settings, particularly in generating faithful rationales \cite{xu2023mental}. Similarly, MentalGLM improved explanation generation through supervised fine-tuning on mental health-specific data \cite{zhai2024mentalglm}. Yet, few studies have proposed structured frameworks to quantify the alignment between model explanations and clinical reasoning. Our work addresses this gap through a span- and symptom-level evaluation framework that enables systematic measurement of explanation \textit{faithfulness} and diagnostic relevance.

\subsection{Prompting Strategies for Mental Health Tasks}

Prompt-based learning, especially in the form of in-context learning (ICL), has proven effective for adapting LLMs to specialized domains. Previous work has shown the utility of prompting for classification and emotion detection in mental health contexts \cite{shi2024mentalqlm}. Inspired by these findings, we design zero-shot and few-shot prompts tailored to symptom identification and rationale generation, leveraging state-of-the-art proprietary LLMs such as GPT-4.1, Gemini 2.5 Pro, and Claude 3.7 Sonnet. Unlike prior works that use weak supervision or synthetic data, our prompting approach is grounded in a gold-standard dataset annotated by clinical experts.

\subsection{Comparative Evaluation of LLMs}

Comparative analyses have shed light on the trade-offs between proprietary and open-source LLMs in mental health applications. Studies like Mental-LLM highlight how instruction tuning can boost model alignment with mental health tasks \cite{xu2023mental}, while other efforts have explored cost-efficient adaptation strategies such as LoRA \cite{shi2024mentalqlm}. In our work, we benchmark leading proprietary models using consistent prompting strategies on our expert-annotated dataset. This allows us to evaluate differences in explanation quality, faithfulness, and symptom coverage, offering insights into how well current LLMs align with clinical reasoning under realistic zero and few-shot conditions.

\section{Methodology}
\label{method}
Our methodology is designed to evaluate the capability of large language models (LLMs) to generate faithful, clinically meaningful natural language explanations (NLEs) for depressive content in social media posts. We leverage our expert-annotated dataset to probe models under prompting-based setups and assess their outputs using a structured evaluation framework.

\subsection{Model Setup and Prompting Strategies}
\label{methodology prompts}
We experiment with state-of-the-art proprietary LLMs, including GPT-4.1, Gemini 2.5 Pro, and Claude 3.7 Sonnet, accessed via APIs. To generate explanations, we employ two prompting strategies:

\paragraph{Zero-Shot Prompting.} In this setup, models are given only a task description and the input post. The prompt instructs the model to identify depressive cues and explain its reasoning without any examples. (Appendix \ref{Appendix zero shot prompt})

\paragraph{Few-Shot Prompting.} Here, we provide models with a small set of in-context examples curated from our expert-annotated dataset. Each example includes a post and a corresponding expert-written explanation. These demonstrations guide the model to better align with the desired explanation format and content. (Appendix \ref{appendix few shot prompt})

Prompts are manually crafted and iteratively refined to ensure consistency and domain alignment across models. For each input post, we collect model-generated explanations under both prompting setups.

\subsection{Evaluation Framework} \label{eval_framework}
To assess explanation quality, we propose a multi-faceted evaluation framework grounded in clinical reasoning and expert feedback. Our evaluation comprises three components:

\paragraph{1. Span-Level Faithfulness Evaluation.}  
We measure how well the model's explanation identifies the same text spans as the expert annotators. This is done via span-level Recall checking if the expert annotated depressive spans are part of explanation or not. We prioritize recall to assess whether models capture the complete range of clinically significant indicators identified by experts. This evaluates the faithfulness of the explanation to the evidence in the input post.

\paragraph{2. Symptom-Level Alignment Evaluation.}  
We extract symptom mentions from the explanation and compare them with the annotated symptom labels using F1-score. This quantifies whether the explanation aligns with clinically grounded categories, reflecting its diagnostic utility and balancing both precision and recall in symptom identification.

\paragraph{3. LLM-Based Coherence Scoring.}  
We use an LLM to rate each explanation based on clarity, logical structure, and emotional interpretation. The prompt is as follows:

\begin{quote}
\texttt{
Given the explanation below for a social media post related to depression, rate it on a scale of 1 to 5 based on the following:
\begin{itemize}
    \item Does it accurately interpret the emotional content of the post?
    \item Is the explanation logically coherent and well-structured?
    \item Does it clearly and correctly tie phrases in the post to depressive symptoms?
    \item Is the explanation concise, avoiding unnecessary repetition or irrelevant elaboration?
\end{itemize}
Return a single rating from 1 (poor) to 5 (excellent). \\
Post: "..." \\
Explanation: "..."
}
\end{quote}

We normalize the score to a 0--1 scale for aggregation.

\paragraph{Combined Quality Metric.}  
To enable holistic comparison of explanation quality across models, we define a unified \textit{Explanation Quality Score} as a weighted sum of three evaluation dimensions:

\begin{equation}
\begin{split}
\mathbf{Quality} &= \mathbf{w_1} \cdot \mathbf{\text{Span-Recall}} + \mathbf{w_2} \cdot \mathbf{\text{Symptom-F1}} \\
&+ \mathbf{w_3} \cdot \left( \frac{\text{Coherence Score}}{5} \right)
\end{split}
\end{equation}

Here, $w_1$, $w_2$, and $w_3$ are weights summing to 1, with values set to 0.3, 0.4, and 0.3, respectively. These values were determined in consultation with clinical experts, who prioritized accurate symptom attribution (captured by \textbf{Symptom-F1}) as the most clinically actionable metric. Span-level fidelity and overall coherence were considered equally important but slightly less critical.

This weighting scheme reflects a 70\% emphasis on clinical alignment (span recall and symptom attribution) and 30\% on the structural and communicative clarity of the explanation. All three component metrics are reported independently to ensure interpretability and enable task-specific analysis.

\subsection{Implementation Details}

All model outputs and evaluations are conducted on a per-post basis. For span extraction, we implement a text-matching procedure that checks whether expert-annotated depressive spans from the original post are explicitly referenced in the model's explanation. This is done using exact and fuzzy matching to account for minor paraphrasing while maintaining semantic alignment.

Symptom identification is performed using keyword matching approach, which maps the explanation text to our predefined set of 12 DSM-5 depression symptoms. This ensures robust detection even when the model uses varied clinical language or synonyms.

To ensure consistency in qualitative judgments, the LLM judge (GPT-4.1) is prompted with a fixed, calibrated instruction across all evaluations, reducing prompt-induced variance.


\section{\textsc{Dataset}} \label{dataset}

\begin{figure*}[t]
    \centering
    \includegraphics[width=0.95\textwidth]{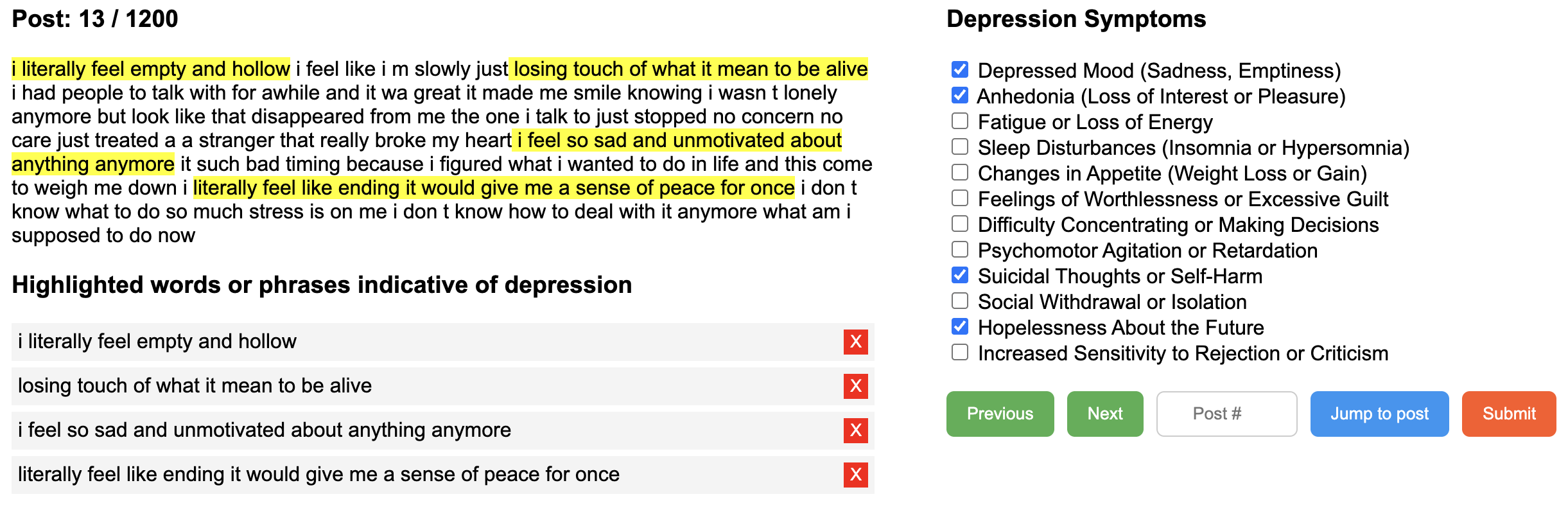}
    \caption{Annotation interface used by certified annotators to highlight depressive spans and select DSM-5-aligned symptom categories.}
    \label{fig:annotation-ui}
\end{figure*}

To support the development of explainable and clinically meaningful depression detection systems, we construct a gold-standard dataset of social media posts annotated by certified mental health professionals. Unlike prior works that rely on weak or crowd-sourced supervision \cite{yates2017depression, coppersmith2015clpsych}, our dataset contains span-level annotations and symptom-level categorizations grounded in DSM-5 diagnostic criteria, allowing for precise and clinically aligned evaluation of language model explanations.

\subsection{Data Collection}

We curated a corpus of publicly available English-language social media posts indicative of emotional distress or self-disclosure. Posts were filtered for quality and relevance using standard preprocessing techniques, including deduplication, language detection, and removal of spam and off-topic content. Posts of sufficient length and semantic clarity were retained to support reliable expert annotation. 

\subsection{Annotation Protocol}

Each post was annotated by certified clinical psychologists and licensed mental health professionals. The annotation interface (Figure~\ref{fig:annotation-ui}) allowed annotators to:

\begin{itemize}
    \item \textbf{Highlight spans} of text expressing depressive thoughts, emotions, or behaviors.
    \item \textbf{Select all depressive symptoms} present in the post from the following 12 DSM-5-derived categories \cite{apa2013dsm}:
    \begin{enumerate}
        \item Depressed Mood (Sadness, Emptiness)
        \item Anhedonia (Loss of Interest or Pleasure)
        \item Fatigue or Loss of Energy
        \item Sleep Disturbances (Insomnia or Hypersomnia)
        \item Changes in Appetite (Weight Loss or Gain)
        \item Feelings of Worthlessness or Excessive Guilt
        \item Difficulty Concentrating or Making Decisions
        \item Psychomotor Agitation or Retardation
        \item Suicidal Thoughts or Self-Harm
        \item Social Withdrawal or Isolation
        \item Hopelessness About the Future
        \item Increased Sensitivity to Rejection or Criticism
    \end{enumerate}
\end{itemize}

Annotations were stored in a structured JSON format, linking each post with its depressive spans and the corresponding symptom labels. The annotation process was governed by standardized guidelines, ensuring consistency and clinical validity across entries. \ref{lst:json-annotation}

\begin{lstlisting}[language=json, 
    caption={Example of annotated post with depressive spans and symptom labels stored in JSON format.},
    captionpos=b,      % places caption at the bottom
    label={lst:json-annotation},
    basicstyle=\ttfamily\scriptsize,
    breaklines=true,
    breakindent=0pt,
    postbreak=\mbox{\textcolor{gray}{$\hookrightarrow$}\space},
    frame=single,
    columns=flexible,
    float=t,
    abovecaptionskip=1ex,
    belowcaptionskip=0pt
]
{
  "14": {
    "post": [
      "i hate myself so much for being ................."
    ],
    "words": [
      "i hate myself so much",
      "i just feel so bitter",
      "they don't deserve me",
      "i'm garbage"
    ],
    "symptoms": [
      0,  // Depressed Mood
      5   // Feelings of Worthlessness or Excessive Guilt
    ]
  }
}
\end{lstlisting}

\subsection{Annotation Guidelines}

To ensure reliable and clinically valid annotations, we developed a comprehensive set of guidelines:

\begin{itemize}
    \item \textbf{Span Selection Criteria}: Annotators were instructed to highlight text spans that directly express or strongly imply a DSM-5 depression symptom. Spans should be:
    \begin{itemize}
        \item Self-contained enough to convey the depressive indicator.
        \item Inclusive of contextual markers necessary for interpretation.
        \item Minimal in length while capturing the complete depressive expression.
    \end{itemize}
    
    \item \textbf{Symptom Classification Guide}: For each symptom category, annotators received detailed descriptions with example phrases and contextual indicators. For instance:
    \begin{itemize}
        \item For "Depressed Mood," annotators looked for expressions of persistent sadness, emptiness, or emotional pain.
        \item For "Anhedonia," they identified statements about diminished interest or pleasure in activities once enjoyed
        \item For "Suicidal Thoughts or Self-Harm," they noted both explicit mentions and indirect references to death wishes, self-harm, or suicide.
    \end{itemize}

\end{itemize}

\subsection{Dataset Statistics}

Our final dataset includes:
\begin{itemize}
    \item \textbf{1017} expert-annotated social media posts.
    \item \textbf{Average post length:} approximately 90.6 words (range: 20-180 words).
    \item \textbf{Symptom distribution:} 
    \begin{itemize}
        \item Most frequent: Depressed Mood (55.8\% of posts), Suicidal Thoughts or Self-Harm (56.5\%), Hopelessness About the Future (35.2\%).
        \item Least frequent: Changes in Appetite (3.4\%), Increased sensitivity to Rejection or Criticism (2\%).
        \item Average of 2.4 distinct symptoms per post annotated
    \end{itemize}
    \item \textbf{Symptom coverage:} All 12 categories are represented, ensuring rich symptom diversity.
\end{itemize}

\subsection{Applications and Impact}

This dataset can have various applications such as:
\begin{itemize}
    \item \textbf{Fine-tuning open-source LLMs} for symptom-aware depression detection and explanation.
    \item \textbf{Faithfulness evaluation} of LLM-generated explanations by comparing generated spans against expert highlights.
    \item \textbf{Multi-level interpretability research}, enabling both token-level and symptom-level assessments.
    \item \textbf{Clinical validation} of AI-based mental health screening tools.
    \item \textbf{Developing responsible disclosure protocols} for depression detection systems.
\end{itemize}

To the best of our knowledge, this is the first dataset to combine span-level depressive indicators with DSM-5-aligned symptom labels. While the dataset comprises 1,017 posts, its strength lies in the depth and precision of annotation rather than sheer volume. Each post is annotated by licensed mental health professionals a process that is inherently time-consuming, cognitively demanding, and requires clinical expertise. This level of expert involvement makes large-scale annotation costly and less feasible, but ensures high-quality, clinically meaningful data. Consequently, even with a modest size, the dataset offers rich supervision and is well-suited for evaluating model faithfulness and interpretability in sensitive applications like mental health detection.


\section{Experiments} \label{experiments}

We design our experiments to evaluate the ability of large language models (LLMs) to generate faithful, clinically relevant explanations for depressive social media posts. All experiments are conducted using prompting-only setups across multiple proprietary LLMs, leveraging the expert-annotated dataset introduced in Section~\ref{dataset}.

\subsection{Models Evaluated}

We assess the performance of the following state-of-the-art proprietary LLMs via API:

\begin{itemize}
\item GPT-4.1 (OpenAI)
\item Claude 3.7 Sonnet (Anthropic)
\item Gemini 2.5 Pro (Google DeepMind)
\end{itemize}

Each model is evaluated under two prompting conditions: \textbf{zero-shot} and \textbf{few-shot}.

\subsection{Prompting Strategies}

To identify the most clinically faithful prompting approaches, we experimented with four variations in the \textbf{Zero-Shot} setting and three in the \textbf{Few-Shot} setting. These prompts varied in terms of clarity, specificity, and the degree to which they emphasized alignment with standard clinical symptoms of depression.

Each prompt’s outputs were reviewed in collaboration with clinical experts to assess their relevance, coherence, and alignment with expert reasoning. Based on this qualitative feedback, the most effective prompt from each category was selected for final evaluation.

For the few-shot setting, we used a fixed set of synthetically constructed examples, each corresponding to one of the 12 standard depressive symptoms (Section~\ref{dataset}). These examples were designed to ensure broad and balanced symptom coverage without overlapping with the evaluation set. Detailed prompts and examples are provided in Appendix~\ref{appendix prompts}.

\subsection{Dataset Usage}

All 1,017 expert-annotated posts are used for evaluation. Importantly, the few-shot prompting examples are not drawn from this dataset; instead, they are created independently to avoid overlap and data leakage. This setup ensures a clear separation between example inputs and evaluation instances.

\subsection{Implementation Details}

All model outputs are generated via the OpenRouter API. To promote faithfulness and reduce variance, inference is conducted with a temperature setting of 0.7. Output truncation is applied as needed to fit within each model’s context window (e.g., 1K tokens for GPT-4.1, 1.5K tokens for Gemni 2.5 pro, 1K for Claude 3.7 Sonnet).

\section{Results and Analysis} \label{results_analysis}

\begin{table*}[h]
\centering
\begin{tabular}{lccccc}
\toprule
\textbf{Model} & \textbf{Setting} & \textbf{Span Eval} & \textbf{Symptom Eval} & \textbf{Coherence Score} & \textbf{Combined Rating} \\
\midrule
GPT-4.1 & Zero-shot & \textbf{0.618} & 0.546 & 4.932 & \textbf{0.6997} \\
Claude 3.7 & Zero-shot & 0.464 & \textbf{0.588} & \textbf{4.936} & 0.6705 \\
\midrule
GPT-4.1 & Few-shot & \textbf{0.578} & 0.587 & \textbf{4.987} & 0.7074 \\
Gemini 2.5 pro & Few-shot & 0.569 & \textbf{0.626} & 4.934 & \textbf{0.7171} \\
Claude 3.7 & Few-shot & 0.454 & 0.592 & 4.952 &  0.6701 \\
\bottomrule
\end{tabular}
\caption{Comprehensive evaluation of models across prompting settings. Combined rating is the weighted sum of span, symptom, and coherence scores with weights $w_1 = 0.3$, $w_2 = 0.4$, and $w_3 = 0.3$. (refer \ref{eval_framework})}
\label{tab:combined-results}
\end{table*}

\subsection{Comparative Model Performance}

Table~\ref{tab:combined-results} presents an evaluation of three flagship language models across zero-shot and few-shot settings. Our framework assessed models on span-level alignment with expert annotations, symptom identification F1 score, and coherence of generated explanations, combining these metrics into an overall performance rating.

\subsubsection{Zero-Shot Performance}

In zero-shot settings, GPT-4.1 demonstrated superior overall performance (combined rating: 0.6997), primarily due to its exceptional span evaluation score (0.618). Claude 3.7 achieved a lower span score (0.464) but outperformed in symptom identification (0.546 vs. 0.588). Gemini 2.5 Pro was excluded from zero-shot analysis as its outputs consistently failed to conform to our expected format despite multiple prompt variations.

\subsubsection{Few-Shot vs. Zero-Shot Comparison}

Contrary to expectations, few-shot prompting did not lead to consistent performance improvements across models. While GPT-4.1 exhibited a modest increase in overall combined rating—from 0.6997 to 0.7074 Claude 3.7 saw a slight decrease, dropping from 0.6705 to 0.6701. The span detection metrics were particularly revealing: GPT-4.1’s span score declined notably from 0.618 to 0.578, and Claude's span performance remained relatively low across both settings.

These results suggest that while few-shot prompting enhances coherence and symptom-level identification, it may inadvertently hinder span detection. One plausible explanation is that the in-context examples anchor the model’s attention to stylistic patterns or conceptual structures, rather than encouraging comprehensive span extraction. Notably, the explanations generated under zero-shot settings were often longer and more exploratory in nature, potentially enabling broader coverage of depressive expressions. This hypothesis is supported by the qualitative differences observed in (Appendix~\ref{appendix:outputs}), where zero-shot responses tend to include a wider array of symptom-related phrases.

\subsubsection{Symptom Identification and Coherence}

Symptom identification remained relatively stable across prompting settings. Claude 3.7 achieved the highest symptom evaluation in the zero-shot setting (0.588), while Gemini 2.5 Pro outperformed others in the few-shot setting (0.626). Coherence scores were consistently high across all models, ranging from 4.932 to 4.987 on a 5-point scale, suggesting that the generated explanations were generally fluent and well-structured regardless of prompting strategy.

\subsection{Model-Specific Analysis}

\subsubsection{GPT-4.1}

GPT-4.1 demonstrated the highest overall performance in the zero-shot setting, achieving the top span evaluation score (0.618) and the highest combined rating (0.6997). This indicates a strong inherent ability to detect depressive cues without the need for in-context examples. Interestingly, its performance improved slightly in the few-shot setting on coherence and symptom dimensions but saw a drop in span alignment, suggesting that exemplars may have inadvertently narrowed its focus. This pattern reflects GPT-4.1's sensitivity to prompt structure and highlights the trade-off between guided and unguided generation in clinical explanation tasks.

\subsubsection{Claude 3.7}

Claude 3.7 demonstrated the highest zero-shot symptom identification score (0.588) but consistently underperformed in span evaluation across both settings. This indicates that while the model is adept at capturing the overall presence of depressive symptoms, it struggles to accurately pinpoint specific span-level indicators within the text. The drop in performance with few-shot prompting suggests that structured examples may have further limited its flexibility in identifying diverse depressive expressions.

\subsubsection{Gemini 2.5 Pro}

Gemini 2.5 Pro was evaluated exclusively in the few-shot setting, as its zero-shot outputs exhibited significant formatting inconsistencies and deviated from the expected structure required for reliable annotation alignment. Specifically, the model often failed to produce clearly delineated symptom explanations or misinterpreted the prompt intent entirely, making span and symptom extraction unreliable in the zero-shot case.

In the few-shot setting, however, Gemini achieved the highest combined rating (0.7171) across all evaluated models, indicating strong overall alignment with expert annotations. It showed balanced performance in both span detection (0.569) and symptom identification (0.626). These results suggest that Gemini benefits substantially from structured prompting with domain-adapted examples, which likely help it organize its outputs more coherently and attend to clinically meaningful cues. When provided with clear guidance, the model is capable of producing detailed, contextually appropriate explanations that support both reasoning transparency and annotation fidelity.

\subsection{Implications for Clinical Applications}

The observed variation in model performance across evaluation dimensions underscores the complexity of automating mental health assessment. Although all models produced generally coherent and readable explanations, their alignment with expert annotations particularly in span detection and symptom classification varied significantly.

GPT-4.1’s relatively strong zero-shot performance highlights its potential utility in real-time or low-resource clinical settings, where prompt engineering and extensive tuning may not be feasible. However, the inconsistent gains from few-shot prompting suggest that curated examples do not uniformly enhance clinical fidelity and must be rigorously validated through collaboration with domain experts.

The trade-off between span-level precision and accurate symptom attribution indicates that no single model currently offers comprehensive reliability across tasks. A hybrid system—using GPT-4.1 for identifying depressive spans and Claude 3.7 or Gemini 2.5 Pro for symptom interpretation may provide a more balanced and clinically useful solution. This approach parallels how clinicians synthesize surface-level cues with deeper diagnostic reasoning. For a qualitative comparison of model explanations, see (Appendix~\ref{appendix comparative remarks}).

\section{Conclusion \& Future Work}

In this work, we introduce a clinically grounded, expert-annotated dataset for explainable depression symptom detection from social media posts. Each post is labeled with fine-grained depressive spans and corresponding symptom categories based on DSM-5 criteria, establishing a gold-standard resource for evaluating model interpretability and clinical relevance.

To complement the dataset, we propose a multi-dimensional evaluation framework that scores explanations across three axes: fidelity to annotated spans, correctness of symptom attribution, and structural coherence, the latter assessed by a consistent LLM-based rubric. This enables a nuanced understanding of how well generative models align with clinical reasoning, beyond simple classification accuracy.

Through comparative experiments using zero-shot and few-shot prompting, we analyze the behavior of state-of-the-art language models under different instruction regimes. Our findings underscore both the promise and limitations of LLMs in generating structured, symptom-aware explanations, with implications for their use in low-resource or high-stakes clinical contexts.

Looking forward, we also aim to refine prompting techniques to encourage conciseness and reduce hallucinations, as well as explore hybrid prompting–fine-tuning pipelines. Ultimately, we hope this resource will serve as a foundation for training and evaluating more interpretable, clinically aligned, and deployable mental health AI systems.




\section*{Limitations}

While our work introduces a high-quality, expert-annotated dataset and a comprehensive evaluation framework, several limitations remain:

\begin{itemize}
    \item \textbf{Dataset size:} Our dataset consists of 1,017 annotated posts, which, although curated carefully, may not fully capture the linguistic and cultural diversity present in real-world mental health discourse. This may limit the generalizability of model evaluations.

    \item \textbf{Subjectivity in annotation:} Despite relying on clinical experts, the span-level and symptom annotations are inherently subjective. Language expressing psychological distress is often ambiguous and open to interpretation, which can lead to variability.

    \item \textbf{LLM-based evaluation bias:} We use large language models to assess the structural coherence and faithfulness of explanations. While scalable, these models may introduce biases or fail to capture clinically subtle or incorrect reasoning that human raters might detect.

    \item \textbf{Limited scope of model evaluation:} Our experiments are restricted to zero-shot and few-shot prompting with proprietary LLMs. We do not evaluate fine-tuned open-source models, which may perform differently and offer stronger clinical alignment in future studies.
\end{itemize}



\section*{Ethical Considerations}

Given the sensitive nature of mental health data and the implications of automated analysis, we took several measures to ensure ethical integrity throughout this work.

\begin{itemize}
    \item \textbf{Data anonymization and privacy:} All social media posts used in this study were publicly available with any personally identifiable information (PII) removed prior to annotation. We ensured compliance with platform terms of service and respected user privacy.

    \item \textbf{Expert annotation and clinical caution:} Annotations were conducted by mental health professionals to promote clinical reliability. However, our dataset is intended strictly for research purposes and should not be used for clinical diagnosis or intervention without human oversight.

    \item \textbf{Model misuse and risk:} Large language models trained or evaluated on this dataset could be misused to make automated decisions about individuals' mental health. We emphasize that such models should not be deployed in real-world settings without rigorous validation, ethical review, and expert involvement.

    \item \textbf{Bias and representation:} Depression manifests differently across cultures, languages, and demographic groups. Our dataset, though carefully curated, may not reflect the full range of experiences, and downstream models may inherit these biases if not used critically.
\end{itemize}

\bibliography{anthology,custom}

\appendix

\section{Appendix: Prompting Techniques for Depression Detection}

\subsection{Prompting Techniques}
\label{appendix prompts}

We employed two main prompting approaches in our experiments: zero-shot and few-shot prompting. Both approaches instructed the model to determine whether a social media post showed signs of depression and generate an explanation by identifying phrases indicative of depression and mapping them to standard symptom categories. (section \ref{methodology prompts})

\subsubsection{Zero-Shot Prompt}
\label{Appendix zero shot prompt}

Our zero-shot prompting approach did not provide any examples but instructed the model with the following prompt:

\begin{small}
\begin{tcolorbox}[breakable, enhanced]
You are a mental health assistant. Determine whether the following post shows signs of depression. If it does, generate explaination by considering phrases from post indicative of depression and depression symptoms present in post from the following categories of standared symptoms.\\
\\
Symptom categories: \\
\\
Depressed Mood (Sadness, Emptiness)

Anhedonia (Loss of Interest or Pleasure)

Fatigue or Loss of Energy

Sleep Disturbances (Insomnia or Hypersomnia)

Changes in Appetite (Weight Loss or Gain)

Feelings of Worthlessness or Excessive Guilt

Difficulty Concentrating or Making Decisions

Psychomotor Agitation or Retardation

Suicidal Thoughts or Self-Harm

Social Withdrawal or Isolation

Hopelessness About the Future

Increased Sensitivity to Rejection or Criticism\\
\\
This is the imput post,\\
\\
Post: "{post}"\\
\\
Explanation:\\
\end{tcolorbox}
\end{small}

\subsubsection{Few-Shot Prompt}
\label{appendix few shot prompt}

Our few-shot prompting approach extended the zero-shot prompt by adding:
\begin{enumerate}
    \item Detailed definitions with examples for each symptom category
    \item Four exemplar posts with expert-crafted explanations
\end{enumerate}

The few-shot prompt structure was:

\begin{small}
\begin{tcolorbox}[breakable, enhanced]
You are a mental health assistant. Determine whether the following post shows signs of depression. If it does, generate explaination by considering phrases from post indicative of depression and depression symptoms present in post from the following categories of standard symptoms.\\
\\
Symptom categories: [list of 12 symptom categories (same as zero shot refer \ref{Appendix zero shot prompt})]\\
\\
here are definitions with example for each symptom category: [detailed symptom definitions and examples refer \ref{Appendix symptom category definitions}]\\
\\
these are some few shot examples for the task:\\
\\
Post: "I don't feel like doing anything anymore. I just wake up feeling exhausted and don't have the energy to get out of bed. My friends keep inviting me to hang out, but I can't seem to find any joy in it. Everything feels pointless, and I don't know why I even bother."\\
\\
Explanation: This post indicates several depressive symptoms:\\
- **Anhedonia**: "I don’t feel like doing anything anymore" and "I can’t seem to find any joy in it" suggest a lack of interest in activities that would normally be enjoyable.\\
- **Fatigue or Loss of Energy**: "I just wake up feeling exhausted and don’t have the energy to get out of bed" points to fatigue and loss of energy, common in depression.\\
- **Depressed Mood**:  "Everything feels pointless" shows a sense of emptiness, which often accompanies sadness in depression.\\
The post expresses a deep lack of motivation, emotional exhaustion, and disconnection from enjoyable activities and social life. These are strong indicators of depressive symptoms.\\
\\

[3 additional example post-explanation pairs]\\
\\
This is the imput post,\\
Post: "{post}"\\
Explanation:
\end{tcolorbox}
\end{small}

\subsubsection{Symptom Category Definitions}
\label{Appendix symptom category definitions}

For the few-shot approach, we provided comprehensive definitions and examples for each symptom category. An excerpt from these definitions includes:

\begin{small}
\begin{tcolorbox}[breakable, enhanced]
1. Depressed Mood (Sadness, Hopelessness, Emptiness)\\
Persistent feelings of sadness, despair, or emotional numbness. The individual may feel overwhelmed by negative emotions, leading to thoughts of hopelessness about the future.\\
Example Phrases: "I feel empty inside," "Nothing makes sense anymore," "I just want to disappear."

2. Anhedonia (Loss of Interest or Pleasure)\\
A marked reduction in interest or enjoyment in activities that were once pleasurable, including hobbies, social interactions, or daily tasks. This can lead to withdrawal from personal and professional commitments.\\
Example Phrases: "I don't enjoy anything anymore," "Even my favorite things feel pointless."

[definitions continue for all 12 symptom categories]
\end{tcolorbox}
\end{small}

\section{Model-Generated Explanations}
\label{appendix:outputs}

To qualitatively illustrate model behavior under different prompting strategies, we present example explanations generated by various LLMs for an input from our dataset. The outputs below are grouped by prompting method and model.

Each example includes:
\begin{itemize}
    \item The original social media post (input).
    \item The expert-annotated depressive spans and symptom labels.
    \item The explanation generated by the model.
\end{itemize}

\subsection{Input social media post:}

\texttt{i used to be highly functional before but it now i can barely function at all i take everything just to keep myself alive i go day without eating how am i ever going to get work done sometime i think i wa not meant for this world there is something very wrong with my brain how am i supposed to fix it how do you keep on living like this forever} 
\\
\textbf{Expert annotations for Input:} \\
\\
\textbf{Spans}:\\
\\
"now i can barely function at all",\\
      "i go day without eating",\\
      "not meant for this world",\\
      "something very wrong with my brain",\\
      "how do you keep on living like this" \\
\\      
\textbf{Symptoms}:  \\
\\
Changes in Appetite (Weight Loss or Gain)\\
Feelings of Worthlessness or Excessive Guilt\\
Difficulty Concentrating or Making Decisions\\
Suicidal Thoughts or Self-Harm\\
Hopelessness About the Future\\
\\
\subsubsection{Zero-Shot Prompting}
\textbf{Explanation using GPT-4.1:} \\
\\
Yes, this post shows signs of depression.
\\
**Explanation:**
\\
Several phrases in the post indicate the presence of symptoms from multiple standard categories of depression:
\\
1. **Fatigue or Loss of Energy:**  
   - "i can barely function at all"  
   - "i take everything just to keep myself alive"  
   These phrases suggest significant fatigue and a struggle to maintain basic functioning.
\\
2. **Changes in Appetite (Weight Loss or Gain):**  
   - "i go day without eating"  
   Indicates a loss of appetite and possibly significant weight loss.
\\
3. **Feelings of Worthlessness or Excessive Guilt:**  
   - "there is something very wrong with my brain"  
   This phrase reflects self-blame and a sense of personal defectiveness.
\\
4. **Hopelessness About the Future:**  
   - "how am i ever going to get work done"  
   - "how do you keep on living like this forever"  
   These statements suggest hopelessness and despair about improvement or the future.
\\
5. **Suicidal Thoughts or Self-Harm:**  
   - "sometime i think i wa not meant for this world"  
   This phrase indicates passive suicidal ideation or thoughts of not wanting to exist.
\\
6. **Anhedonia (Loss of Interest or Pleasure):**  
   - "i used to be highly functional before but it now i can barely function at all"  
   Implies a loss of previous abilities and interests.
\\
**Summary:**  
The post contains clear expressions of fatigue, appetite disturbance, hopelessness, feelings of worthlessness, possible suicidal ideation, and a marked decline in functioning and interest—symptoms consistent with depression.
 \\
\\
\textbf{Explanation using Claude 3.7 Sonnet}\\
\\
\# Analysis of Depression Signs in the Post
\\
This post shows significant signs of depression. Here's an analysis of the concerning elements:
\\
\#\# Phrases Indicative of Depression
\\
1. "I used to be highly functional before but now I can barely function at all"\\
2. "I take everything just to keep myself alive"\\
3. "I go day without eating"\\
4. "Sometimes I think I was not meant for this world"\\
5. "There is something very wrong with my brain"\\
6. "How am I supposed to fix it"\\
7. "How do you keep on living like this forever"\\
\\
\#\# Depression Symptoms Present
\\
1. **Fatigue or Loss of Energy**: The statement about being "highly functional before" but now "can barely function at all" indicates severe energy depletion and difficulty performing daily tasks.
\\
2. **Changes in Appetite**: "I go day without eating" directly references disturbed eating patterns, suggesting significant appetite changes.
\\
3. **Feelings of Worthlessness**: The belief that "I was not meant for this world" and that something is "very wrong with my brain" suggests deep feelings of being fundamentally flawed.
\\
4. **Hopelessness About the Future**: The question "how do you keep on living like this forever" demonstrates a bleak outlook and inability to see improvement in their condition.
\\
5. **Difficulty Functioning**: The concern about "how am I ever going to get work done" shows impairment in daily responsibilities.
\\
6. **Possible Suicidal Ideation**: While not explicitly stating suicidal intent, phrases like "not meant for this world" and questioning how to "keep on living like this forever" suggest passive suicidal ideation or existential despair.
\\
The post presents multiple core symptoms of clinical depression, including significant functional impairment, disturbed eating, hopelessness, and possible passive suicidal ideation. The individual appears to be in considerable distress and would likely benefit from professional mental health support.

\subsubsection{Few-Shot Prompting (with Domain-Adapted Examples)}

\textbf{Explaination generated by GPT-4.1}\\
\\
This post shows multiple signs of depression:\\
\\
- **Anhedonia (Loss of Interest or Pleasure):** "i used to be highly functional before but it now i can barely function at all" indicates a marked loss of ability and motivation to engage in life, suggesting the person no longer finds meaning or pleasure in activities they once managed easily.\\
- **Changes in Appetite:** "i go day without eating" is a clear sign of reduced appetite, which is a common symptom in depression and can lead to weight loss and further health complications.\\
- **Feelings of Worthlessness or Excessive Guilt:** "there is something very wrong with my brain" and "i was not meant for this world" express deep self-blame, perceived defectiveness, and a sense of not belonging. These are strong indications of worthlessness and potentially excessive guilt.\\
- **Hopelessness About the Future:** "how am i ever going to get work done" and "how do you keep on living like this forever" reflect a lack of hope for improvement and a belief that the current suffering will persist indefinitely.\\
- **Suicidal Thoughts or Self-Harm:** "sometime i think i wa not meant for this world" suggests thoughts of not belonging or of wishing to no longer exist, which can be an indicator of suicidal ideation.\\
\\
The post conveys a profound loss of function and energy, neglect of basic needs like eating, deep self-criticism, despair about the future, and thoughts of not belonging in the world. These are strong indicators of severe depressive symptoms and possible suicidal ideation, highlighting the need for immediate support and intervention. \\

\textbf{Explaination generated by Gemni 2.5 pro}\\
\\
Yes, this post shows significant signs consistent with depression:\\
\\
-   **Fatigue or Loss of Energy / Psychomotor Retardation**: "i used to be highly functional before but it now i can barely function at all" indicates a severe decline in the ability to engage in daily activities and a potential slowing down, suggesting profound fatigue or psychomotor retardation.\\
-   **Changes in Appetite**: "i go day without eating" points directly to a significant decrease in appetite, a common symptom.\\
-   **Feelings of Worthlessness**: "sometime i think i wa not meant for this world" and "there is something very wrong with my brain" express deep feelings of inadequacy, feeling fundamentally flawed, and not belonging.\\
-   **Hopelessness About the Future**: Phrases like "how am i ever going to get work done", "how am i supposed to fix it", and "how do you keep on living like this forever" convey a strong sense of despair and a belief that the situation is permanent and unfixable.\\
-   **Difficulty Concentrating or Making Decisions**: The concern "how am i ever going to get work done" suggests difficulty managing tasks, which can stem from problems with concentration or decision-making.\\
-   **Suicidal Thoughts**: "sometime i think i wa not meant for this world" can be interpreted as passive suicidal ideation, a serious symptom of depression reflecting a desire to not exist.\\
-   **Depressed Mood**: The overall tone and the final question "how do you keep on living like this forever" strongly suggest feelings of despair, sadness, and hopelessness.
\\
The post describes a drastic loss of function, changes in basic needs like eating, profound hopelessness, feelings of worth\\
\\

\textbf{Explaination generated by Claude 3.7}\\
\\
\# Depression Analysis\\
\\
This post shows significant signs of depression with multiple concerning symptoms:
\\
\#\# Symptoms Present:
\\
- **Fatigue or Loss of Energy**: "i used to be highly functional before but now i can barely function at all" indicates a severe decline in energy and functioning ability compared to previous capability.
\\
- **Changes in Appetite**: "i go day without eating" clearly indicates disturbed eating patterns and significant appetite reduction.
\\
- **Difficulty Concentrating or Making Decisions**: "how am i ever going to get work done" suggests struggles with focus and completing tasks.
\\
- **Feelings of Worthlessness**: "i think i was not meant for this world" reflects profound feelings of not belonging and worthlessness.
\\
- **Hopelessness About the Future**: "how do you keep on living like this forever" demonstrates a bleak outlook and inability to see improvement in the future.
\\
- **Depressed Mood**: The overall tone reflects deep sadness and emptiness, particularly when comparing current state to previous functioning.
\\
- **Possible Suicidal Ideation**: While not explicit, phrases like "i take everything just to keep myself alive" and "i was not meant for this world" suggest thoughts about existence and mortality that could indicate passive suicidal ideation.\\
\\
The post reveals someone experiencing a significant decline in functioning, struggling with basic self-care (eating), feeling fundamentally flawed ("something very wrong with my brain"), and questioning their ability to continue living in this state. These are serious depressive symptoms that suggest the need for professional mental health intervention.

\subsubsection{Comparative Remarks}
\label{appendix comparative remarks}

Across the different prompting strategies and models, we observe both commonalities and key differences in the structure, depth, and clinical alignment of generated explanations.

\textbf{1. Symptom Coverage and Specificity:}  
All models identified core depressive symptoms such as appetite changes, fatigue, hopelessness, and suicidal ideation. However, few-shot prompting generally led to more comprehensive and symptomatically accurate explanations. Models under few-shot settings, particularly GPT-4.1 and Gemini 2.5 Pro, often surfaced more nuanced symptoms like difficulty concentrating and psychomotor retardation, which were sometimes missed in zero-shot outputs.

\textbf{2. Language and Coherence:}  
Claude 3.7 Sonnet produced notably structured and readable explanations even under zero-shot prompting, organizing symptoms into clearly labeled sections. GPT-4.1 and Gemini 2.5 Pro, under few-shot prompting, provided similarly well-organized outputs but with deeper clinical reasoning and contextual interpretation.

\textbf{3. Clinical Alignment with Expert Annotations:}  
Few-shot GPT-4.1 and Gemini 2.5 Pro exhibited the highest alignment with expert-annotated spans and symptom categories. They not only matched key phrases but also interpreted their clinical implications with high fidelity. In contrast, zero-shot models sometimes generalized or omitted less overt symptoms.

\textbf{4. Interpretability and Justification Style:}  
Few-shot prompted explanations were more reflective and interpretive, often providing psychological context (e.g., linking functional decline to anhedonia or existential despair to passive suicidal ideation). Zero-shot explanations tended to be more checklist-like or declarative, lacking explanatory depth.

\textbf{5. Sensitivity to Passive Suicidal Ideation:}  
Most models, especially under few-shot prompting, identified subtle cues such as "not meant for this world" as indicative of suicidal ideation. Claude and Gemini were particularly careful in framing such phrases within a clinical safety context, reflecting greater sensitivity.

\textbf{Conclusion:}  
Few-shot prompting significantly enhances the quality, depth, and clinical validity of model explanations. Among the evaluated models, GPT-4.1 and Gemini 2.5 Pro demonstrated the most faithful alignment with human expert reasoning, both in symptom attribution and narrative interpretation. This underscores the value of prompt engineering when deploying LLMs in sensitive domains like mental health.

\section{Technical Implementation}

All prompts were implemented using the OpenRouter API to access flagship language models. The implementation included temperature settings of 0.7 and a maximum token limit of 1,000 tokens for model responses to ensure detailed but concise explanations.

\begin{lstlisting}[language=Python, 
  caption={Prompting implementation using the GPT-4.1 model.}, 
  label={lst:prompt-impl}, 
  captionpos=b,       % Place caption at the bottom
  basicstyle=\ttfamily\small, 
  breaklines=true, 
  columns=fullflexible
]
def generate_explanation(post):
    prompt = build_prompt(post)
    response = client.chat.completions.create(
        model="gpt-4.1",
        messages=[
            {"role": "user", "content": prompt}
        ],
        temperature=0.7,
        max_tokens=1000)
    return response.choices[0].message.content
\end{lstlisting}

\vspace{0.5em}  

\end{document}